\documentclass[
twocolumn,
]{ceurart}

\usepackage{listings}
\usepackage{tcolorbox}
\usepackage{enumitem}
\begin{document}

\copyrightyear{2022}
\copyrightclause{Copyright for this paper by its authors.
  Use permitted under Creative Commons License Attribution 4.0
  International (CC BY 4.0).}

\conference{AIMLAI'22: In Proceedings of Advances in
Interpretable Machine Learning and Artificial Intelligence (AIMLAI) at CIKM'22}

\title{Privacy and  Transparency in Graph Machine Learning: A Unified Perspective }

\author{Megha Khosla}[%
orcid=0000-0002-0319-3181,
email=m.khosla@tudelft.nl,
url=https://khosla.github.io/,
]
\address{Delft University of Technology, Delft, The Netherlands}

\begin{abstract}
  Graph Machine Learning (GraphML), whereby classical machine learning is generalized to irregular graph domains, has enjoyed a recent renaissance, leading to a dizzying array of models and their applications in several domains. With its growing applicability to sensitive domains and regulations by governmental agencies for trustworthy AI systems,  researchers have started looking into the issues of transparency and privacy of graph learning.
  However, these topics have been mainly investigated independently. In this position paper, we provide a unified perspective on the interplay of privacy and transparency in GraphML. In particular, we describe the challenges and possible research directions for a formal investigation of privacy-transparency tradeoffs in GraphML.
\end{abstract}


\maketitle

\section{Introduction}

Graphs are a highly informative, flexible, and natural way to represent data. 
Graph based machine learning (GraphML), whereby classical machine learning is generalized to irregular graph domains, has enjoyed a recent renaissance, leading to a dizzying array of models and their applications in several fields \cite{10.1093/bib/bbab159, dong2022mucomid,ying2018graph,sanchez2020learning,MPM}. 
GraphML models have achieved great success due to their ability to flexibly learn from the complex interplay of graph structure and node attributes/features. 
Such ability comes with a  compromise in privacy and transparency, two indispensable ingredients to achieve trustworthy ML \cite{dai2022comprehensive}.

Deep models trained on graph data are inherently blackbox, and their decisions are difficult for humans to understand and interpret. The growing application of these models in sensitive applications like healthcare and finance and the regulations by various AI governance frameworks necessitate the need for transparency in their decision-making process. Meanwhile, recent research \cite{iyiola2021,duddu2020quantifying,zhanggraphmi,he2021stealing} has highlighted the privacy risks of deploying models trained on graph data.
It has been suggested that these models are even more vulnerable to privacy leakage than models trained on non-graph data due to the additional encoding of relational structure in the model itself \cite{iyiola2021}.

Consequently, an increasing number of works are focussing on explaining \cite{GNNExplain,Zorro21,luo2020parameterized,vu2020pgm} the decisions of black box GraphML models in a post-hoc manner, designing interpretable models \cite{yu2020graph,rathee2021learnt,zhang2021protgnn} as well as privacy preserving techniques for real world deployments of graph models  \cite{olatunji2021releasing,sajadmanesh2021locally,sajadmanesh2022gap}.  
\begin{figure*}[t!]
    \includegraphics[width=1\textwidth]{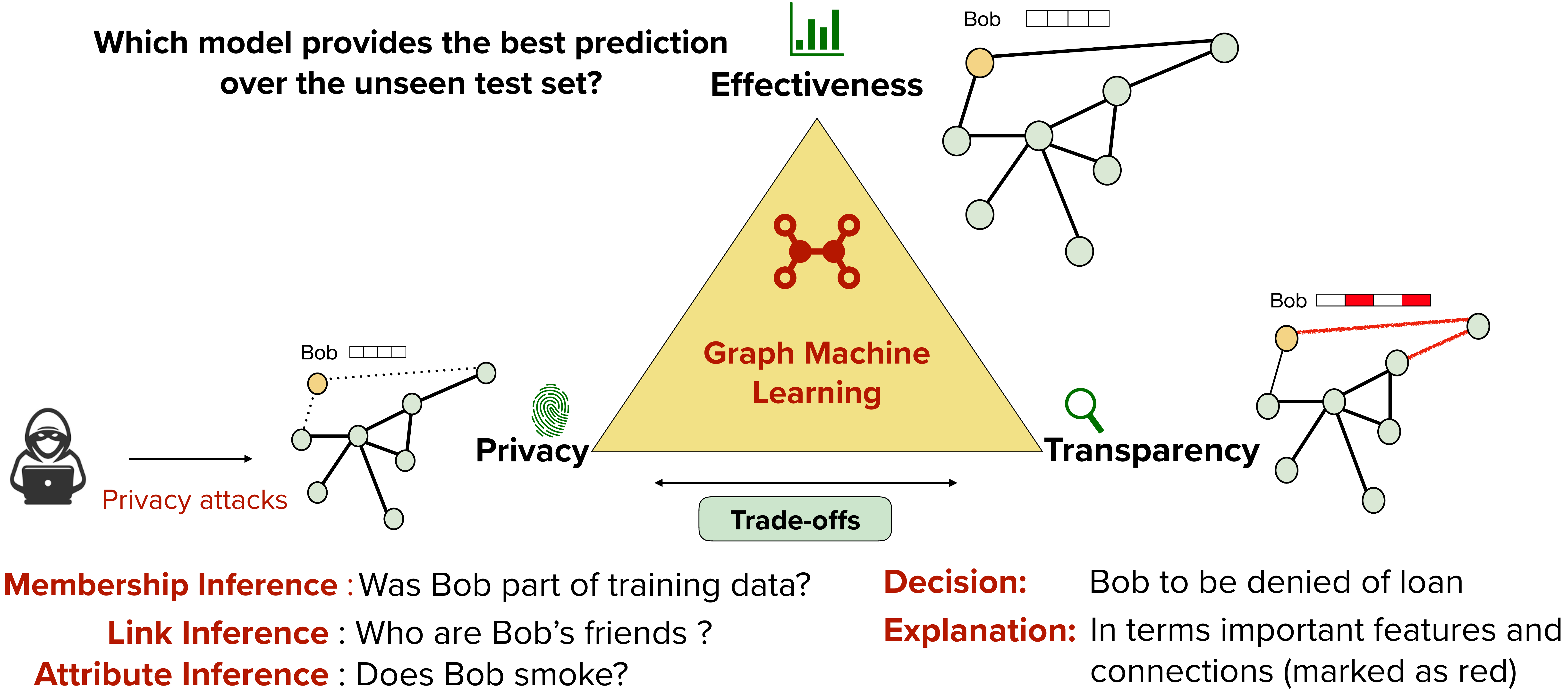}
    \caption{Privacy and transparency  are usually studied together with their effect on model performance. But the trade-offs between privacy and transparency have been so far ignored. \textit{Can transparency increase the risk of privacy leakage? How transparent are privacy preserving models?}}
    \label{fig:overview}
\end{figure*}

Despite the growing research interest, the current state of the art considers privacy and transparency in GraphML independently. While transparency provides insight into the model's working, privacy aims to preserve the sensitive information about the training data\footnote{Here we are only concerned with data privacy. Model Privacy or protecting the model itself against, for example, stealing model parameters is out of the scope of this paper.}. The seemingly conflicting goals of privacy and transparency call for the need of a joint investigation. To date, any gain in privacy or transparency is usually compared to any drop in model performance. However, questions like \textit{``what effects would be releasing post-hoc explanations have on the privacy of training data?''} or \textit{``how well can we interpret the decisions of privacy-preserving graph models?''} have so far received little attention \cite{shokriprivacyexplanations,olatunji2022}.

In this position paper, we provide a unified perspective on the inextricable link between privacy and transparency for GraphML. 
Besides, we sketch the possible research directions towards formally exploring privacy-transparency tradeoffs in GraphML.

\section{Background}
\subsection{Graph Machine Learning} The key idea in graph machine learning is to encode the discrete graph structure into low dimensional continuous vector representations using non-linear dimensionality reduction techniques. Popular classes of GraphML methods include \textit{random walk based} strategies \cite{perozzi2014deepwalk,NERD} which encode structural similarity of the nodes exposed by their co-occurrence in random walks; \textit{matrix-factorization based} \cite{zhou2017scalable}  which rely on low rank factorization of some node similarity matrix; and the most popular \textit{graph neural networks} (GNNs) \cite{kipf2017semi,hamilton2017inductive} which learns node representations by recursive aggregation and transformation of neighborhood features. These methods are usually non-transparent and are shown to be prone to privacy leakage risks.

\begin{tcolorbox}
    Towards improving the adoption of these methods in sensitivity applications like healthcare and medicine the community has started paying attention to the aspects of transparency and privacy. However these aspects have been so far studied independently (see also Figure \ref{fig:overview} for an illustration). A formal investigation into the  linked role of transparency and privacy in achieving trustworthy GraphML is missing.
\end{tcolorbox}
%Despite their popularity in analyzing interconnected data, these approaches are not easy to interpret and are prone to data privacy risks. 

\subsection{Transparency for GraphML Models} 
\begin{figure*}[h!]
    \centering
    \includegraphics[width=0.8\textwidth]{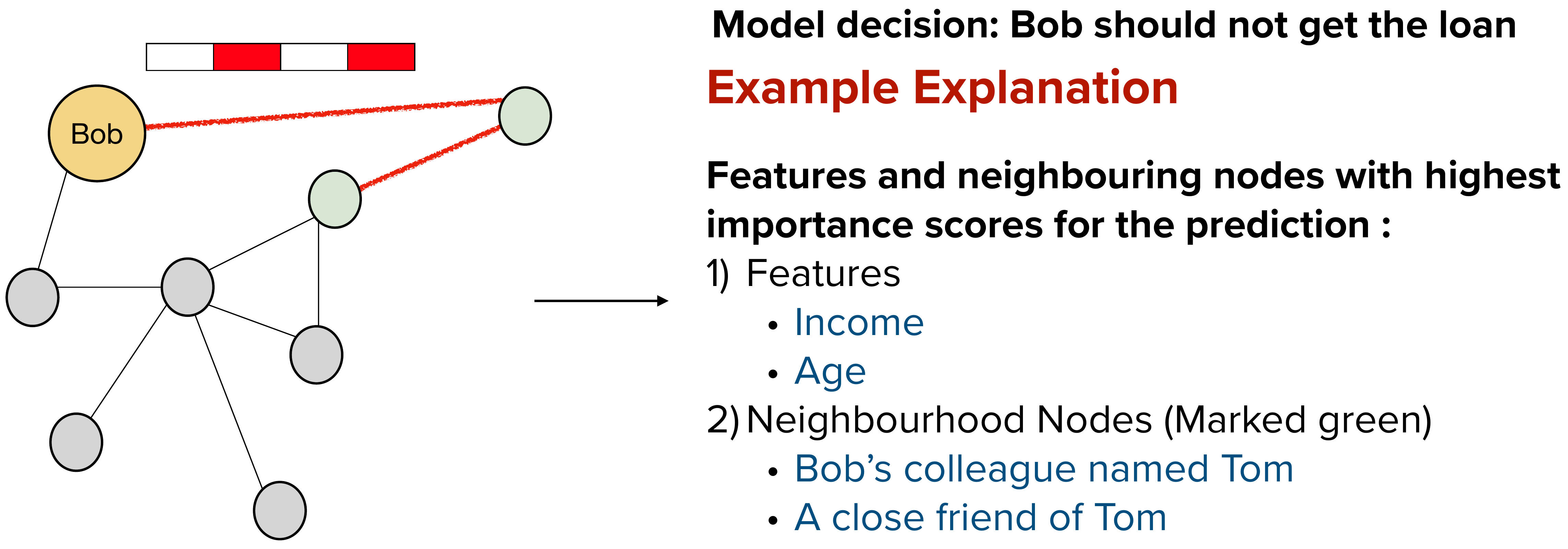}
    \caption{An example explanation in terms of features and node attribution over a  social network in which a node represents a user and edges represent friendship relation. Node features correspond to demographic attributes of the user. Neighboring nodes with high importance scores are marked green.
  }
    \label{fig:explanation}
\end{figure*}
Transparency for deep models, as in GraphML, is usually achieved by providing \textit{explanations} corresponding to decisions of an already trained model or by building \textit{interpretable by design} or \textit{self-explaining} models. Numerous approaches have been proposed in the literature for explaining general machine learning models \cite{chen2018:l2x,yoon2018invase,binder2016:lrp,sundararajan2017:integratedgrad}; however, models learned over graph-structured data have some unique challenges. 

Specifically, predictions on graphs are induced by a complex combination of nodes and paths of edges between them in addition to the node features. A trivial application of existing explainability methods to graph models cannot account for the role of graph structure in the model decision. Consequently several graph specific explainability approaches have been recently developed which focus primarily on explaining graph neural networks' decisions for node and graph classification \cite{XGNN2020,gao2021gnes}. 

Explanations usually include the importance scores for nodes/edges in a subgraph (or node's neighborhood in case of node-level task) and the node features \cite{GNNExplain,Zorro21,luo2020parameterized}. Figure \ref{fig:explanation} depicts an example of an explanation over graph data. Depending on the explanation method, the importance scores could be either continuous (soft masks) or binary (hard masks). A few works have also been proposed to explain dense unsupervised node representations \cite{kang2019explaine,IdahlKA19}.
In terms of methodologies, several techniques based on input perturbations \cite{GNNExplain,Zorro21,luo2020parameterized}, input gradients\cite{pope2019explainability,sanchez2020evaluating}, causal techniques \cite{kang2019explaine,bajaj2021robust,gao2021gnes} as well as utilizing simpler surrogate models \cite{vu2020pgm} have been explored. 

Another methodology to provide transparency is to develop interpretable by design models \cite{yu2020graph,rathee2021learnt,miao2022interpretable}. Such models usually contain a self-explanatory module trained jointly with the learner module. Explanations are thus, by design, faithful to the model. 

A few other works also focus on unifying diverse notions of evaluation strategies \cite{bagel,sanchez2020evaluating} necessary for effectively assessing the quality and utility of explanations.
\begin{tcolorbox}Despite the progress in improving transparency of GraphML techniques its effect on data privacy has escaped attention. While transparency could increase the utility of the model, for sensitive applications any unaddressed concerns for privacy can hinder the full adoption of the models and further dissuade the participants to share their data.
\end{tcolorbox}

\subsection{Privacy in GraphML}
\begin{figure*}
    \centering
 \includegraphics[width=0.7\textwidth]{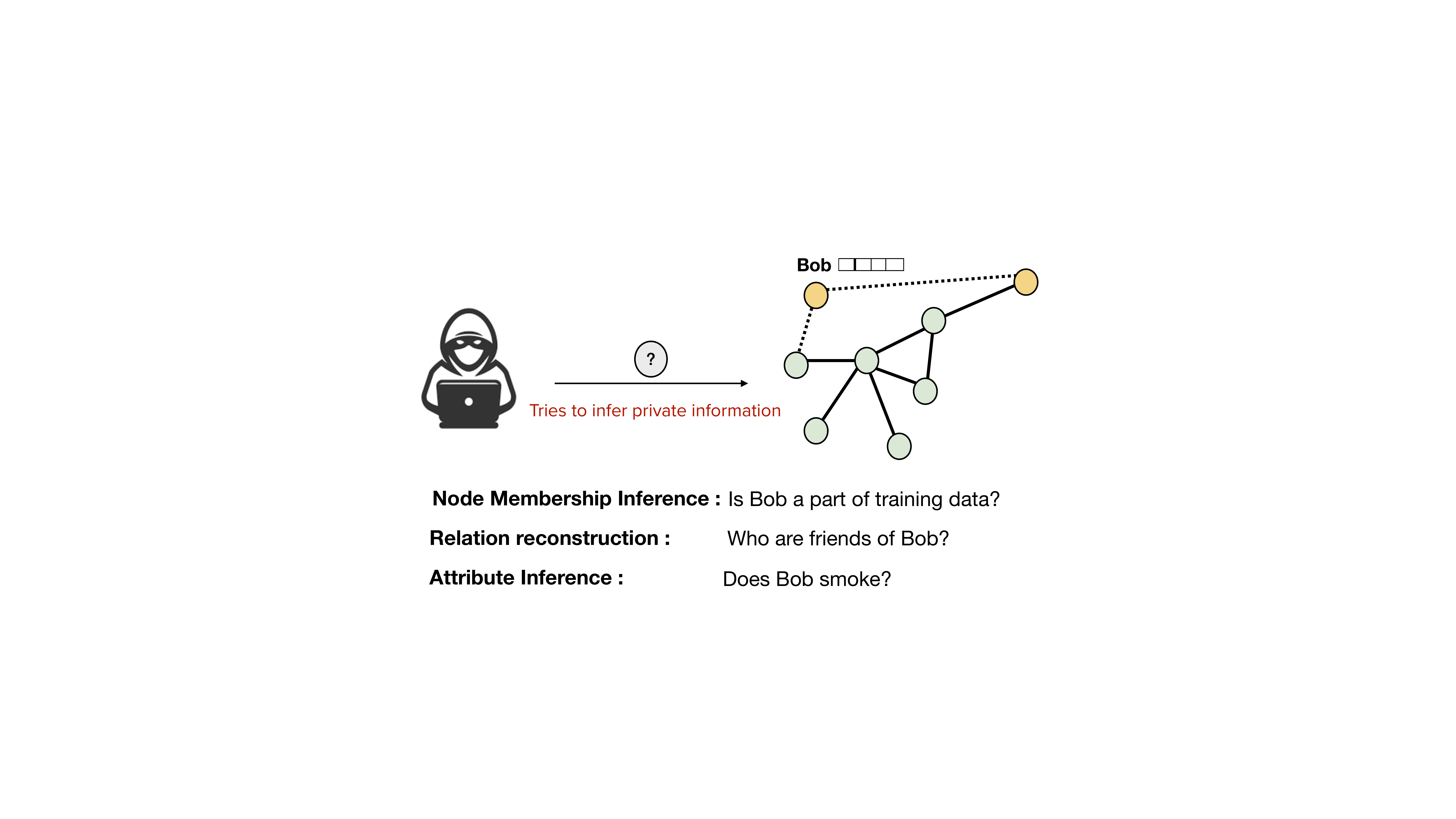}
    \caption{Given access to trained model or embeddings trained on graph data, an adversary can launch several attacks to infer membership, relations or attributes of a node.}
    \label{fig:privacy}
\end{figure*}

Deep learning models, in general, are known to leak private information about the employed training data. 
 Recent works showed that  trained model on graph data can leak sensitive information about the training data (see Figure \ref{fig:privacy}) like node membership \cite{iyiola2021,duddu2020quantifying}, certain dataset properties \cite{zhang2022inference} and connectivity structure of the nodes \cite{zhanggraphmi}. In Figure \ref{fig:privacy} we illustrate the possibility of different privacy attacks given access to trained GraphML model.
 Compared to general deep learning models, GraphML is more vulnerable to privacy risks as they incorporate not only the node features/labels but also the graph structure \cite{iyiola2021}.
 
 Privacy-preserving techniques for graph models are mainly based on differential privacy \cite{xu2018dpne,iyiola2021,sajadmanesh2021locally,sajadmanesh2022gap} and adversarial training frameworks \cite{netfense,liao2021information,li2020adversarial}. The key idea in differential privacy \cite{dwork2006calibrating} is to conceal the presence of a single individual in the dataset. In particular, if we query a dataset containing $N$ individuals, the query's result will be probabilistically indistinguishable from the result of querying a neighboring dataset with one less or one more individual. For machine learning models, such probabilistic indistuinguishability is achieved by adding appropriate levels of noise at different levels of model development. For instance, \cite{xu2018dpne} employs objective perturbation mechanism to develop differential private network embeddings. Olatunji et al. \cite{iyiola2021} combines the knowledge-distillation framework with the
two noise mechanisms, random subsampling, and noisy labeling to release graph neural networks under differential privacy guarantees. In particular it uses only a random sample of private data to train teacher models corresponding to nodes in an unlabelled public dataset. The final model which is later released is trained on public data using the noisy labels generated by the teacher models. Other works \cite{sajadmanesh2022gap, sajadmanesh2021locally} do not build a separate public model but achieve DP via adding noise directly to the aggregation module of GNNs. Adversarial defence to privacy attacks on GNNs is proposed in \cite{netfense}, in which the predictability of private labels is destroyed and the utility of perturbed graphs is maintained. An adversarial learning approach based on mini-max game between the desired graph feature encoder and the worst-case attacker is proposed in \cite{liao2021information} to address the attribute inference attack on GNNs.

\begin{tcolorbox}Despite the growing number of works in improving privacy in GraphML, its effect on transparency of these models is not at all studied. The complex mechanisms employed to ensure privacy further hurts the model transparency. Consequently it is not clear if existing explainers can be used to explain the decision making process of privacy-preserving models.
\end{tcolorbox}
 
\section{A Unified Perspective}

Graphs are powerful abstractions that facilitate leveraging data interconnection to represent, predict, and explain real-world phenomena. Exploiting such explicit or latent data interconnections, on the one hand, makes GraphML more powerful but also brings in additional challenges, further exacerbating the need for a joint investigation of privacy and transparency. In following we discuss the key issues arising due to the independent treatment of privacy and transparency for GraphML.
\subsection{Diverse explanation types and methods} 
Model explanations for graph data are usually in the form of feature and neighborhood (subgraph) attributions. In particular, importance scores for node features and its neighboring nodes/edges are released as explanations. Neighborhood attributions or structure explanations are a more direct form of information leakage. They can be, for example, leveraged to identify nodes in the training set or infer hidden attributes of sensitive nodes using the attributes of their neighbors.

Besides, the data points (nodes) in graph data are correlated, thus violating the usual i.i.d. assumption over data distributions. Consequently, the decisions and explanations over correlated nodes might themselves be correlated. Such correlations among released explanations might be exploited to reconstruct sensitive information of the training data. 
For example, the similarity in feature explanations for recommendations to two connected users might reveal the sensitive link information they want to hide. 
Towards this \cite{olatunji2022} show that the link structure of the training graph can be reconstructed with a high success rate even if only the feature explanations are available.

\subsection{Transparency of private models}
Moreover, due to the correlated nature of the graph data, privacy-preserving mechanisms on graph models need to focus on several aspects such as node privacy, edge privacy, and attribute privacy \cite{sajadmanesh2022gap}. This leads to more complex privacy-preserving mechanisms, which results in a further loss of transparency. To understand the issue, consider a simple differential privacy-based mechanism in which randomized noise is added to the model's output. Such noise could alter the final decision but not the decision process that an explanation (according to its current definition) is usually expected to reveal. Model agnostic approaches for explainability, which only assume black-box access to the trained model, might be misguided by such alteration in the final decision.

\subsection{The curse of overfitting}
 In traditional machine learning, we can randomly divide the data into two parts to obtain training and test sets. It is more tricky in graphs where the data points are connected, and random data sampling may result in non i.i.d. train and test sets. Even for the task of graph classification where the graphs constitute the datapoints instead of the the nodes, distributional changes are common in train and test splits \cite{li2022out} due to varying graph structure and size.
 Specifically, the train set may contain specific spurious correlations which are not representative of the entire dataset. This puts GraphML models at a higher risk of overfitting to sample specific correlations rather than learning the desired general patterns \cite{zhu2021shift}. Existing privacy attacks have leveraged overfitting to reveal sensitive information about the training sample \cite{yeom2020overfitting}. Exploiting associated explanations, which in principle should reveal learned spurious correlations, can further aid in privacy leakage. 

\section{Research Directions}
 Based on the described issues and challenges in the previous section, we recommend the following research directions towards a formal investigation of privacy-transparency tradeoffs.
 \begin{enumerate}[leftmargin=*]\setlength{\itemsep}{5pt}
     \item \textbf{New Threat Models.}  A first step is to quantify the privacy risks of releasing post-hoc explanations. Towards that, we need to design new threat models and \textit{structure-aware} privacy attacks in the presence of  post-hoc model explanations. Care should be taken to formulate \textit{realistic assumptions on adversary's background knowledge.} For example, in highly homophilic graphs, an adversary might be able to approximate well the link structure of the graph only if the node features/labels are available. \textit{What information explanations could leak in addition when explanations are provided?}
     \item \textbf{Risk-utilty assessment of different explanation types and methods.} 
   Model explanations for GraphML can be in the form of feature or node/edge importance scores. Besides, existing explanation methods are based on different methodologies and might be discovering different aspects of the model decision process. Depending on the dataset and application, certain types of explanation methods and types of explanation (feature or structural) might be preferred over others. A dataset and application-specific risk-utility assessment might reveal more favorable explanations for minimizing privacy loss. For instance,  \cite{olatunji2022} finds that gradient-based feature explanations have the least predictive (faithfulness to the model) power for the task of node classification but leak the most amount of information about the private structure of the training graph. In such cases, one can decide not to reveal such an explanation as it has little utility for the user.
   \item \textbf{Transparency of privacy-preserving models.}
Besides evaluating the privacy risks of releasing explanations, it is essential to analyze the transparency of privacy-preserving techniques. It is not clear if existing explanation strategies can faithfully explain the privacy-preserving models' decisions. Questions like  \textit{what should be the properties of explanations of such models? What constitutes a faithful explanation?} need to be investigated. Consequently new techniques to explain privacy preserving models need to be developed.
      \item{\textbf{Reducing overfitting.}} Overfitting is usually considered a common enemy for model effectiveness on unseen data and privacy. 
      Recently, a few works have proposed interpretable by design models for example using stochastic attention mechanisms \cite{miao2022interpretable}, graph sparsification strategies \cite{rathee2021learnt} etc. These methods are claimed to remove spurious correlations in the training phase leading to a reduction in overfitting. A possible research direction is further exploiting such transparency strategies to minimize privacy leakage. 
    \end{enumerate}

\section{Conclusion}
There has been an unprecedented rise in the popularity of graph machine learning in recent years. With its growing applications in sensitive areas, several works focus independently on their transparency and privacy aspects. We provide a unified perspective on the need for a joint investigation of privacy and transparency in GraphML. We hope to start a discussion and foster future research in quantifying and resolving the privacy-transparency tradeoffs in GraphML. Resolution of such tradeoffs would make GraphML more accessible to stakeholders currently tied down by regulatory concerns and lack of trust in the solutions. 

\bibliography{sample-ceur}

\end{document}